\documentclass{llncs}
\usepackage{epsf}

\usepackage[utf8]{vietnam}

\AtBeginDocument{\renewcommand{\abstractname}{Abstract.}}
\AtBeginDocument{\renewcommand{\refname}{References}}
\AtBeginDocument{\renewcommand{\tablename}{Table}}

\begin{document}

\title{A Feature-Rich Vietnamese Named-Entity Recognition Model}

\author{ }
\institute{ }
                
\author{
Pham Quang Nhat Minh}
\institute{Alt Vietnam Co., Ltd\\
			  92 Trieu Viet Vuong, Hanoi, Vietnam\\
                pham.minh@alt.ai}

\maketitle

\begin{abstract}

In this paper, we present a feature-based named-entity recognition (NER) model that achieves the start-of-the-art accuracy for Vietnamese language. We combine word, word-shape features, PoS, chunk, Brown-cluster-based features, and word-embedding-based features in the Conditional Random Fields (CRF) model. We also explore the effects of word segmentation, PoS tagging, and chunking results of many popular Vietnamese NLP toolkits on the accuracy of the proposed feature-based NER model. Up to now, our work is the first work that systematically performs an extrinsic evaluation of basic Vietnamese NLP toolkits on the downstream NER task. Experimental results show that while automatically-generated word segmentation is useful, PoS and chunking information generated by Vietnamese NLP tools does not show their benefits for the proposed feature-based NER model.

\end{abstract}

\section{Introduction}

Named-entity recognition (NER) is an important task in information extraction. The task is to identify in a text, spans that are entities and classify them into pre-defined categories. There have been some conferences and shared tasks for evaluating NER systems in English and other languages, such as MUC-6~\cite{Sundheim1995OverviewOR}, CoNLL 2002~\cite{Sang2002IntroductionTT} and CoNLL 2003~\cite{Sang2003IntroductionTT}.

In Vietnamese language, VLSP 2016~\cite{Huyen2016} is the first evaluation campaign that aims to systematically compare NER systems for Vietnamese language. Similar to CoNLL 2003 shared-task, in VLSP 2016, four named-entity types were considered: person(PER), organization (ORG), location (LOC), and miscellaneous entities (MISC). NER systems in VLSP 2016 adopted either conventional feature-based sequence labeling models such as Conditional Random Fields (CRFs), Maximum-Entropy-Markov Models (MEMMs) or recurrent neural network (RNN) with LSTM units. The first rank NER system in VLSP 2016 applied MEMMs with specific features for Vietnamese NER data~\cite{Le2016VietnameseNE}.

In this paper, we formalize NER task as a sequence-labeling problem and propose a feature-rich NER model for Vietnamese NER, which use word, word-shape features, PoS tags, chunking tags, and features based on two types of word representations: Brown word clusters and word embedding. We adopt CRF~\cite{Lafferty:2001}, a popular sequence-labeling method for our NER model. On the first data set of VLSP NER evaluation with provided word segmentation, PoS, and chunking tags, our system obtained the state-of-the-art $F_1$ score. Our proposed system significantly outperforms previous work on Vietnamese NER, including a more complicated NER model, which combines bidirectional Long Short-Term Memory (Bi-LSTM), Convolutional Neural Network (CNN), and Conditional Random Fields~\cite{pham2017nnvlp}.

There are two NER data sets provided in VLSP 2016 campaign. While the first data set contains word segmentation, PoS, chunking, named entity (NE) information, the second dataset contains only NE information. In the first data set, word segmentation is gold-standard word segmentation. Although PoS tags and chunking tags were generated automatically by public tools, they were partly corrected by annotators during the annotation process~\footnote{We obtained that information thanks to an online discussion with a member in VLSP 2016 organizers}. In the overview paper~\cite{Huyen2016}, there is no mention about tools which the VLSP 2016 organizer used to determine PoS and chunking tags.

To date, many published work on Vietnamese NER has reported evaluation results on the first data set. They have used default word segmentation, PoS, and chunking tags provided by organizers of VLSP 2016. However, we could not obtain word segmentation, PoS and chunking tags that way for NER in real scenarios. There is no work that explored the effects of automatically generated word segmentation, PoS, and chunking tags on the accuracy of Vietnamese NER models. Our work will fill that gap by comparing the usage of automatically generated word segmentation, PoS, and chunking tags generated by popular off-the-self Vietnamese NLP toolkits in NER task. Experimental results show that while automatically-generated word-segmentation is useful for a feature-based NER model, PoS and chunking information generated by Vietnamese NLP tools did not give their benefits.

The remainder of the paper is organized as follows. Section~\ref{sec:related_work} presents some related work to our research. In Section~\ref{sec:proposed_system}, we describe our NER system. Next, in Section~\ref{sec:exp_design}, we present the design of experiments in the paper.  In Section~\ref{sec:result}, we present experimental results achieved on the VLSP 2016 NER data set.  Finally, in Section~\ref{sec:conclusion}, we give conclusions and some remarks.

\section{Related Work}
\label{sec:related_work}

Basically, we can categorize machine-learning approaches to NER into conventional machine-learning models and deep-learning models. Conventional machine-learning methods often adopted models such as Conditional Random Fields~\cite{Lafferty:2001}, Hidden Markov Models, Support Vector Machines, or Maximum-Entropy Markov Models. Those methods require to design hand-crafted features for NER~\cite{florian2003named}. In contrast, deep-learning NER models do not require hand-crafted features but the computational cost in training is very high compared with conventional machine-learning models~\cite{TACL792}.

For Vietnamese, VLSP community has organized the first evaluation campaign for NER in 2016. Vietnamese NER systems that evaluated on the VLSP 2016 data applied either conventional machine-learning or deep-learning methods. The first rank system in the campaign used MEMM and obtained $89.66\%$ $F_1$ score on the test data~\cite{Le2016VietnameseNE}.

Recently, Pham and Le-Hong, 2017~\cite{Pham2017a} incorporated word embedding and syntactic features including PoS, chunk, and regular expressions in Bi-LSTM model and acquired $92.05\%$ $F_1$ score. They claimed that automatic syntactic features improve $F_1$ score about $18\%$. Pham et al., 2017~\cite{pham2017nnvlp} combined Bi-LSTM, CNN, CRF and obtained $92.91\%$ $F_1$ score. We argue that syntactic features they used are not really automatic syntactic features because PoS and chunking tags provided in the NER dataset were partly corrected by annotators during the annotation process.

In the best of our understanding, all published Vietnamese NER papers that used the VLSP 2016 NER dataset reported result on the data with default word-segmentation, PoS, chunking tags provided by the VLSP 2016 organizers. There is no work that investigate the effects of automatically generated word-segmentation, PoS tags, and chunking tags by published Vietnamese NLP toolkits to the downstream NER task. Our paper is the first work that addresses that issue.

\section{Proposed Feature-Based Vietnamese NER Model}
\label{sec:proposed_system}

We formalize NER task as a sequence labeling problem by using the B-I-O tagging scheme and we apply a popular sequence labeling model, Conditional Random Fields to the problem. In this section, we briefly describe CRF, and then present features that we used in our model.

\subsection{Conditional Random Fields}

Conditional Random Fields~\cite{Lafferty:2001} is a discriminative probabilistic framework,
which directly model conditional probabilities of a tag sequence given
a word sequence. Formally, in CRF, the
conditional probability of a tag sequence $y = (y_1, y_2, \dots, 
y_m)$, given a word sequence $x = (x_1, x_2, \dots, x_m)$ is defined as follows.

\begin{equation}
  P(y|x) = \frac{\exp( w \cdot F(y,x))}{\sum_{y' \in Y} \exp( w \cdot F(y',x))}
\end{equation}
where $w$ is the parameter vector to be estimated from
training data; $F(y,x) \in {\rm I\!R}^d$ is a global feature function that is defined on an entire input sequence and an entire tag sequence; $Y$ is the space of all possible tag sequences. The feature function $F(y,x)$ is calculated by summing local feature functions.

\begin{equation}
F_j(y,x) = \sum_{i=1}^n f_j(y_{i-1}, y_i, x, i)
\end{equation}

The parameters in CRF can be estimated by maximizing log-likelihood objective function. Parameter estimation in CRF can be done by using iterative scaling algorithms or gradient-based methods~\cite{Lafferty:2001}.

\subsection{Features}

Basically, features in the proposed NER model are categorized into word, word-shape features, PoS and chunking tag features, features based on word representations including word clusters and word embedding. Note that, we extract unigram and bigram features within the context surrounding the current token with the window size of $5$. More specifically, for a feature $F$ of the current word, unigram and bigram features are as follows.

\begin{itemize}
\item \textbf{unigrams}: $F$[-2], $F$[-1], $F$[0], $F$[1], $F$[2]
\item \textbf{bigrams}: $F$[-2]$F$[-1], $F$[-1]$F$[0], $F$[0]$F$[1], $F$[1]$F$[2]
\end{itemize}

\subsubsection{Word Features}
	
We extract word-identity unigrams and bigrams within the window of size 5. We use both word surfaces and their lower-case forms. Beside words, we also extract prefixes and suffixes of surfaces of words within the context of the current word. In our model, we use prefixes and suffixes of lengths from 1 to 4 characters.

\subsubsection{Word Shapes}

In addition to word identities, we use word shapes to improve prediction ability, especially for unknown or rare words and reduce data spareness problem.

Word shape features are summarized in the Table~\ref{tab:wordForms}. Among word shape features,  we extract both unigram and bigram features for ``shaped'', ``type'', and ``fregex''. For other word shapes, only unigrams are extracted. Features from ``fregex'' to ``wei'' were proposed in~\cite{Le2016VietnameseNE}.

\begin{table}[t]
  \caption{Word shape features}
  \label{tab:wordForms}
  \centering
  \begin{tabular}{|l | l | l |}
    \hline
    \textbf{Feature} & \bf Description & \textbf{Example} \\ 
    \hline  \hline
    shape & orthographic shapes of the token &  ``\textit{Đồng}'' $\rightarrow$ ``\textit{ULLL}''\\
    shaped & shorten version of shape & ``\textit{Đồng}'' $\rightarrow$ ``\textit{UL}''\\
    type & category of the token such as ``AllUpper'', ``AllDigit'', etc & ``1234'' $\rightarrow$  ``AllDigit''\\
    fregex & features based on token regular expression~\cite{Le2016VietnameseNE} & \\
    mix & is mixed case letters & ``\textit{iPhone}''\\
    acr & is capitalized letter with period & ``\textit{H.}'', ``\textit{Th.}'',
                                        ``\textit{U.S.}'\\
    ed & token starts with alphabet chars and ends with digits & ``\textit{A9}'', ``\textit{B52}''\\
    hyp & contains hyphen & ``\textit{New-York}''\\
    da & is date & ``\textit{03-11-1984}'', ``\textit{03/10}''\\
    na & is name & ``\textit{Buôn\_Mê\_Thuột}''\\
    co & is code & ``\textit{21B}''\\
    wei & is weight & ``\textit{2kg}''\\
    2d & is two-digit number & ``\textit{12}''\\
    4d & is four-digit number & ``\textit{1234}''\\
    d\&a & contains digits and alphabet & ``\textit{12B}''\\
    d\&- & contains digits and hyphens & ``\textit{9-2}''\\
    d\&/ & contains digits and backslash & ``\textit{9/2}''\\
    d\&, & contains digits and comma & ``\textit{10,000}''\\
    d\&. & contains digits and period & ``\textit{10.000}''\\
    up & contains an upper-case character followed by a period & ``\textit{M.}''\\
    iu & first character is upper-case & ``\textit{Việt\_Nam}''\\
    au & all character of the token are upper-case & ``\textit{IBM}''\\
    al & all characters are lower-case &  ``\textit{học\_sinh}''\\
    ad & all digits & ``\textit{1234}''\\
    ao & all characters are neither alphabet characters nor digits & ``\textit{;}''\\
    cu & contains at least one upper-case character & ``\textit{iPhone}''\\
    cl & contains at least one lower-case character &  ``\textit{iPhone}''\\
    ca & contains at least one alphabet character & ``\textit{s12456}'' \\
    cd &  contains at least one digit & ``\textit{1A}''\\
    cs & contains at least 1 character that is not alphabet or digit & ``\textit{10.000}''\\
    \hline
  \end{tabular}
\end{table}

\subsubsection{PoS and chunking tags}

Similar to word features, we extract unigrams and bigrams of PoS tags and chunking tags of words within the window of size 5.

\subsubsection{Brown cluster-based features}

Brown clustering algorithm is a hierarchical clustering algorithm for assigning words to clusters~\cite{Brown:1992:CNG:176313.176316}. Each cluster contains words which are semantically similar. Output clusters are represented as bit-strings. In natural language processing, word clusters can be used to tackle the problem of data sparseness by providing lower-dimensional representations of words. The usage of brown-cluster-based features have been explored for named-entity recognition in the work of Miller~\cite{miller-guinness-zamanian:2004:HLTNAACL}, and then widely used in discriminative learning NLP models~\cite{koo-carreras-collins:2008:ACLMain,turian-ratinov-bengio:2010:ACL}.

Brown-cluster-based features in our NER model include whole bit-string representations of words and their prefixes of lengths 4, 6, 8, and 10. Note that, we only extract unigrams for Brown-cluster-based features.

In experiments, we used the Brown clustering implementation of Liang~\cite{liang2005semi} and applied the tool on the raw text data collected through a Vietnamese news portal. We performed word clustering on the same preprocessed text data which were used to generate word embeddings in~\cite{le2017empirical}. The number of word clusters used in our experiments is 1000.

\subsubsection{Word embeddings}

Word-embedding-based features have been used for a CRF-based Vietnamese NER model in~\cite{le2017empirical}. The basic idea is adding unigram features corresponding to dimensions of word representation vectors. 

In the paper, we apply the same word-embedding features as in~\cite{le2017empirical}. We generated pre-trained word vectors by applying Glove~\cite{pennington2014glove} on the same text data used to run Brown clustering. The dimension of word vectors in 25. 

\section{Experimental Design}
\label{sec:exp_design}

\subsection{Dataset}

In experiments, we used the NER data set from VLSP 2016 evaluation campaign with default train/test split. There are 16,858 sentences in training data and 2,381 sentences in test data. The data set contains nested entities, yet we only consider first level entities in this paper. The statistics of the data set is shown in Table~\ref{tab:dataset}.
\begin{table}[t]
  \center
  \caption{Statistics of named entities in the VLSP corpus}
  \label{tab:dataset}
  \begin{tabular}{|l|r|r|}
    \hline 
    \textbf{Entity Types} & \textbf{Training Set} & \textbf{Test Set}
    \\ 
    \hline
    \hline 
    Location & 6,245 & 1,379 \\ 
    \hline 
    Organization & 1,213 & 274 \\ 
    \hline 
    Person & 7,480 & 1,294 \\ 
    \hline 
    Miscellaneous names & 282 & 49 \\ 
    \hline 
    All & 15,222 & 2,996 \\ 
    \hline 
  \end{tabular}
\end{table}

The data set provided by VLSP 2016 organizers contains word-segmentation, PoS, and chunking tags along with NER tags. While word-segmentation is manually annotated by human, PoS and chunking tags were automatically determined by tools and then partly corrected by annotators during annotation process.

\subsection{CRF Tool and Parameters}

In experiments, we adopted CRFsuite~\cite{CRFsuite}, an implementation of linear-chain (first-order Markov) CRF. That toolkit allows us to easily incorporate both binary and numeric features such as word embedding features. In training, we use Stochastic Gradient Descent algorithm with L2 regularization and the coefficient for L2 regularization is $3.2$.

\subsection{Default and Generated PoS, Chunking Tags}

In the VLSP 2016 NER data, PoS and chunking tags were not determined in a fully automatic manner. In our understanding, all published Vietnamese NER work that evaluated on VLSP 2016 data use default word-segmentation, PoS and chunking tags. In real scenarios, we could not obtain PoS and chunking tags that way. In this work, we compare the performance of our NER system in two settings: using default PoS and chunking tags and using PoS and chunking tags generated by off-the-self Vietnamese toolkits. We investigate the effect of PoS, and chunking tags to only our NER model. We plan to do same experiments using other Vietnamese NER models in the future work.

Because of the space limitation, we could not investigate all Vietnamese NLP toolkits in the paper. We choose two Vietnamese toolkits to perform chunking: Underthesea~\footnote{\url{https://github.com/magizbox/underthesea}} and NNVLP~\cite{pham2017nnvlp}. To perform PoS tagging, we use Underthesea, NNVLP, Pyvi~\footnote{\url{https://pypi.python.org/pypi/pyvi}}, Vitk~\footnote{\url{https://github.com/phuonglh/vn.vitk}}, and VnMarMoT~\cite{NguyenVNDJ-ALTA-2017}. Those tools are all popular Vietnamese NLP toolkits. We keep the original word-segmentation when we run Vietnamese PoS and chunking tools on the training and test portions of the NER data to reduce the error propagation from word-segmentation tools.

\subsection{Default and Generated Word Segmentation}

Each word in Vietnamese language may consist of one or more syllables with spaces in between. For instance a location name ``\textit{Hà Nội}'' consists of two syllables ``\textit{Hà}'' and ``\textit{Nội}''. The VLSP 2016 dataset is word segmented, in which spaces between syllables in multi-syllable words were replaced by underscores ``\_''. Because there is no mention about how word segmentation was generated in~\cite{Huyen2016} and organizer reused the dataset for PoS tagged of VLSP project~\footnote{\url{http://vlsp.hpda.vn:8080/demo/?\&lang=en}}, we believe that word segmentation in the VLSP 2016 NER dataset was manually annotated. In this paper, we compare our NER model when we train and test on data with default and generated word segmentation. We also perform an extrinsic evaluation for popular word-segmentation tools in the NER task.

In order to re-generate word segmentation on the training and test data, we remove all word segmentation info in the data, and then run Vietnamese word segmentation tools on the obtained data. We keep the syllables tokenized in the data to avoid boundary-conflict problem in evaluation on the test data segmented by tools. Some tool, such as pyvi tokenizes syllables in the original data into smaller units. For instance ``Mr.'' is tokenized to ``\textit{Mr}'' and ``\textit{.}''. Thus, we choose word segmentation tools that allow us to perform word segmentation on the data with syllables tokenized in advanced. We choose two word segmentation tools, UETSegmenter~\footnote{\url{https://github.com/phongnt570/UETsegmenter}} and RDRsegmenter~\footnote{\url{https://github.com/datquocnguyen/RDRsegmenter}}, which are perfectly fit our need. The two tools obtained good word segmentation results. UETSegmenter obtained 98.82\% $F_1$ score~\cite{nguyen2016hybrid}, and RDRsegmenter obtained 97.90\% $F_1$ score on the benchmark Vietnamese treebank~\cite{NguyenNVDJ2018}.

\subsection{Syllable-Based Model and Word-Based Model}

In this paper, we further investigate the effect word segmentation to the proposed Vietnamese NER model by a comparing syllable-based CRF model with a word-based CRF model. In the syllable-based model, BIO tags are tagged on syllable units. In order to generate training and test data for the syllable-based model, we convert BIO tags of words in the original data to BIO tags for syllables. For instance, in word-based model the location ``\textit{Hà\_Nội}'' is tagged with ``\textit{B-LOC}'' tag, and in syllable-based model, the word will be converted into two syllables with tags: ``\textit{Hà/B-LOC}" and ``\textit{Nội/I-LOC}''. We hypothesize that word-segmentation is useful for NER task and using automatically generated word segmentation improves the accuracy of feature-based NER models against the syllable-based model.

Word embeddings and Brown clusters which we learned for word-based model contained segmented words, so many syllables are not included in vocabularies of them. Therefore, in experiments, we learned word embeddings and Brown clusters for syllable-based model on the unsegmented version of raw text corpora which were used to generate Brown clusters for the word-based model.

\section{Main Results}
\label{sec:result}
\begin{table}[t]
\caption{Accuracy of our NER system with full features set and default PoS and chunking tags}
\label{tab:default_tags}
\begin{center}
\begin{tabular}{lccc}
\hline
\bf System & \bf Precision & \bf Recall & \bf $F_1$\\
\hline
Vitk~\cite{Le2016VietnameseNE} & 89.56 & 89.75 & 89.66\\
vie-ner-lstm~\cite{Pham2017a} & 91.09 & 93.03 & 92.05\\
NNVLP~\cite{pham2017nnvlp} & 92.76 & 93.0 & 92.91\\
\bf Our System & \bf 93.87 & \bf 93.99 & \bf 93.93\\
\hline 
\end{tabular}
\end{center}
\end{table}


Table~\ref{tab:default_tags} shows the accuracy of our NER model and previous NER models using the dataset with default word-segmentation, PoS, chunking tags. In experiments, we use micro-averaged $F_1$ score, the official evaluation metric in CoNLL 2003~\cite{Sang2003IntroductionTT} as the evaluation measure. We compare our NER model with following Vietnamese NER models.

\begin{itemize}
\item Vitk~\cite{Le2016VietnameseNE} is the system that obtained the first rank in the VLSP 2016 evaluation campaign. In that work, authors combines regular expressions over tokens and a bidirectional inference method in a sequence labelling model.
\item vie-ner-lstm~\cite{Pham2017a} incorporates syntactic features including PoS, Chunk and regular-expression-based features into a bidirectional Long Short-Term Memory (Bi-LSTM) model. They claimed that incorporating automatic syntactic features improves $F_1$ score about $18\%$.
\item NNVLP~\cite{pham2017nnvlp} applied Bi-LSTM-CNN-CRF with pre-train word embeddings for Vietnamese language. That model also used default word-segmentation, PoS, chunking tags of VLSP NER dataset.
\end{itemize}

Results in Table~\ref{tab:default_tags} indicated that, our feature-based NER model outperforms the previous work with a large margin. We obtain $93.93\%$ of $F_1$ score on the test set, which is $1\%$ higher than NNVLP system.

\subsection{The Effect of PoS and Chunking Tags}

\begin{table}[t]
\caption{Accuracy of our NER system with default and generated PoS, chunking tags; and without PoS and chunking tags}
\label{tab:autoPoSChunk}
\begin{center}
\begin{tabular}{lccc}
\hline
\bf Setting & \bf Precision & \bf Recall & \bf $F_1$\\
\hline
Default PoS and chunking tags & 93.87 & 93.99 & 93.93\\
PoS and chunking tags generated by NNVLP~\cite{pham2017nnvlp} & 90.21 & 86.72 & 88.43\\
PoS and chunking tags generated by Underthesea & 90.28 & 88.35 & 89.3\\
\bf Without PoS, chunking tags & 89.91 & 90.15 & \bf 90.03\\
\hline 
\end{tabular}
\end{center}
\end{table}

In Table~\ref{tab:autoPoSChunk}, we show experimental results of our system when we apply automatic Vietnamese PoS tagging and chunking tools to generate PoS and chunking tags. We can see that with automatically-generated PoS and chunking tags, $F_1$ score of the system dropped significantly, which is $4.63\%$. Incorporating automatically generated PoS and chunk by tools NNVLP or Underthesea did not improve the accuracy of the NER model. Underthesea tool showed the better result than NNVLP when they were used in our NER model.

A plausible explanation for the result is that chunking tags encode information about boundary of entity mentions. Entities often occur within a noun phrase. Therefore, correct chunking tags will help to improve accuracy of a NER model. 

We observe original chunking tags in VLSP NER data and chunking tags generated by NNVLP and by Underthesea, and see that there is a big gap between the original ones and generated ones. The following example shows original chunking tags and generated ones of a sentence in the training data.
\begin{itemize}
\item \textbf{Original chunking tags}: ``Một/B-NP chuyến/B-NP hải\_trình/B-NP xuyên/B-VP ba/B-NP nước/B-NP Malaysia/B-NP ,/O Singapore/B-NP ,/O Indonesia/B-NP vừa/O được/B-VP phóng\_viên/B-NP Tuổi\_Trẻ/B-NP thực\_hiện/B-VP ,/O''
\item \textbf{By NNVLP}: ``Một/B-NP chuyến/I-NP hải\_trình/I-NP xuyên/B-VP ba/B-NP nước/I-NP Malaysia/I-NP ,/O Singapore/B-NP ,/O Indonesia/B-NP vừa/O được/B-VP phóng\_viên/B-NP Tuổi\_Trẻ/I-NP thực\_hiện/B-VP ,/O''
\item \textbf{By Underthesea}: Một/B-NP chuyến/B-NP hải\_trình/B-NP xuyên/B-VP ba/B-NP nước/I-NP Malaysia/I-NP ,/I-NP Singapore/I-NP ,/I-NP Indonesia/I-NP vừa/B-VP được/I-VP phóng\_viên/B-NP Tuổi\_Trẻ/I-NP thực\_hiện/I-NP ,/O
\end{itemize}

In original chunking tags, ``Malaysia'', ``Singapore'', ``Indonesia'' make three noun phrases. NNVLP tool tagged ``ba nước Malaysia'' (``three countries Malaysia'') as one noun phrase, and Underthesea tagged ``ba nước Malaysia, Singapore, Indonesia'' (``three countries Malaysia, Singapore, Indonesia'') as one noun phrase. Underthesea incorrectly tagged ``phóng\_viên Tuổi\_Trẻ thực\_hiện'' (``done by reporter of Tuoi Tre News'') as a noun phrase. 

The feature-based NER model learns useful patterns from correct chunking tags. Patterns learned from incorrect generated chunking tags even become noises to the machine-learning model.

\begin{table}[t]
\caption{Proposed NER systems without chunking tag-based features. We compare default PoS with PoS generated by other tools.}
\label{tab:autoPoS}
\begin{center}
\begin{tabular}{lccc}
\hline
\bf Setting & \bf Precision & \bf Recall & \bf $F_1$\\
\hline
Default PoS tags & 90.13 & 90.55 & 90.34\\
PoS by NNVLP~\cite{pham2017nnvlp} & 90.05 & 85.65 & 88.31\\
PoS by Underthesea & 90.27 & 88.58 & 89.42\\
PoS by Pyvi & 90.16 & 88.72 & 89.43\\
PoS by Vtik & 89.62 & 86.42 & 87.99\\
PoS by VnMarMoT~\cite{NguyenVNDJ-ALTA-2017} & 90.51 & 89.15 & 89.83\\
\bf Without PoS, chunking tags & 89.91 & 90.15 & \bf 90.03\\
\hline 
\end{tabular}
\end{center}
\end{table}

In the next experiment, we remove chunk features in the model, and compare the accuracy of the model when we use PoS tags generated by different tools.
Table~\ref{tab:autoPoS} indicated that with default PoS tags, our NER model obtained highest $F_1$ score. However incorporating PoS tags generated by other PoS tagging tools did not help to improve against the model without PoS and chunking tags.

Since in VLSP 2016 dataset, PoS tags were not automatically generated, we can safely say that automatically generated PoS tags did not give benefits to our feature-based NER model.

\subsection{The Effect of Word Segmentation}

\begin{table}[t]
\caption{Accuracy of NER system with default and generated word segmentation. We did not use features based on PoS, chunking tags here.}
\label{tab:autows}
\begin{center}
\begin{tabular}{lccc}
\hline
\bf Setting & \bf Precision & \bf Recall & \bf $F_1$\\
\hline
Default Word segmentation & 89.91 & 90.15 & 90.03\\
Word segmentation generated by UETSegmenter & 87.67 & 84.95 & 86.29\\
Word segmentation generated by RDRsegmenter & 89.05 & 84.98 & \bf 86.97 \\
\hline 
\end{tabular}
\end{center}
\end{table}

\begin{table}[t]
\caption{Accuracy of NER system with syllable-based and word-based model. We do not use features based on PoS and chunking tags. ``ws'' stands for word segmentation}
\label{tab:syllable}
\begin{center}
\begin{tabular}{lccc}
\hline
\bf Setting & \bf Precision & \bf Recall & \bf $F_1$\\
\hline
Syllable-based model & 88.78 & 82.94 & 85.76\\
Word-based model with gold ws & 89.91 & 90.15 & 90.03\\
Word-based model with ws generated by RDRsegmenter  & 89.05 & 84.98 & \bf 86.97 \\

\hline 
\end{tabular}
\end{center}
\end{table}

Similarly, Table~\ref{tab:autows} shows comparison of the model accuracy with default word-segmentation and word segmentation generated by the two state-of-the-art Vietnamese segmentation tools: UETSegmenter and RDRsegmenter. Word segmentation result of RDRsegmenter leads to better NER accuracy compared with UETSegmenter.

In comparison with using default word-segmentation (which is manually annotated word-segmentation), the $F_1$ of  score of our model with automatic word segmentation decreased about $3\%$. That suggests that there is still room for improvement of Vietnamese word segmentation, especially in downstream NLP tasks. 

Table~\ref{tab:syllable} show results of syllable-based models and word-based models. Experimental results confirm our hypothesis that in Vietnamese, word segmentation is useful for a feature-based NER model. Word-based models outperform syllable-based models even with automatically generated word-segmentation.

\subsection{The Effect of Word-representation-based Features}

\begin{table}[t]
\caption{Impact of word representation-based features. w2v denotes features based on word embeddings. ``cluster'' denotes cluster-based features.}
\label{tab:wordRepr}
\begin{center}
\begin{tabular}{lccc}
\hline
\bf Setting & \bf Precision & \bf Recall & \bf $F_1$\\
\hline
(1) = all features with default PoS, Chunk & 93.87 & 93.99 & 93.93\\
(2) = (1) - cluster - w2v & 91.66 & 92.02 & 91.84\\
(4) = word + word shapes + default PoS & 88.01 & 87.95 & 87.98\\
(5) = word + word shapes + cluster + w2v & 89.91 & 90.15 & 90.03\\
(6) = word + word-shapes & 88.17 & 88.08 & 88.13\\
(7) = word + word-shapes + w2v & 88.69 & 88.72 & 88.70\\
(8) = word + word-shapes + cluster & 88.96 & 89.99 & 89.97\\
\hline 
\end{tabular}
\end{center}
\end{table}

In order to evaluate the impact of word representation-based features, we conducted experiments with different feature sets. We start with a feature set, then remove features related to word clusters and word vectors. Results in Table~\ref{tab:wordRepr} indicated the importance of word representation-based features. Incorporating those features improves $F_1$ score more than $2\%$. Resuls also showed that Brown cluster-based features contribute more to the system improvement than word-embedding features. An advantage of word-representations is that they can be learned in the unsupervised fashion from raw-text corpora.

\section{Conclusion}
\label{sec:conclusion}

In the paper, we presented a feature-based named-entity recognition model for Vietnamese language, which obtains the state-of-the-art accuracy on the standard VLSP 2016 NER data set. Using default word-segmentation, PoS, chunking tags provided by VLSP 2016 organizers, our system achieved 93.93\% $F_1$ score. We showed that, in our CRF-based NER model, PoS and chunking features are useful if PoS and chunking tags are precise. However automatically generated PoS and chunking tags did not give their benefit to the accuracy improvement. We pointed out that word-segmentation in Vietnamese language is useful to the downstream NER task, and word-segmentation generated by state-of-the-art Vietnamese word segmentation tools is helpful, but that there is still a big gap between the usage of manually annotated word segmentation and that of automatically generated word segmentation in a feature-based NER model.

\bibliographystyle{splncs}
\bibliography{paper}

\begin{thebibliography}{10}

\bibitem{Sundheim1995OverviewOR}
Sundheim, B.:
\newblock Overview of results of the muc-6 evaluation.
\newblock In: MUC. (1995)

\bibitem{Sang2002IntroductionTT}
Sang, E.F.T.K.:
\newblock Introduction to the conll-2002 shared task: Language-independent
  named entity recognition.
\newblock CoRR \textbf{cs.CL/0209010} (2002)

\bibitem{Sang2003IntroductionTT}
Sang, E.F.T.K., Meulder, F.D.:
\newblock Introduction to the conll-2003 shared task: Language-independent
  named entity recognition.
\newblock In: CoNLL. (2003)

\bibitem{Huyen2016}
Huyen, N.T.M., Luong, V.X.:
\newblock Vlsp 2016 shared task: Named entity recognition.
\newblock In: Proceedings of Vietnamese Speech and Language Processing (VLSP).
  (2016)

\bibitem{Le2016VietnameseNE}
Le, H.P.:
\newblock Vietnamese named entity recognition using token regular expressions
  and bidirectional inference.
\newblock CoRR \textbf{abs/1610.05652} (2016)

\bibitem{Lafferty:2001}
Lafferty, J., McCallum, A., Pereira, F.:
\newblock Conditional random fields: {P}robabilistic models for segmenting and
  labeling sequence data.
\newblock In: ICML. (2001)  282--289

\bibitem{pham2017nnvlp}
Pham, H., Khoai, P.X., Nguyen, T.A., Le-Hong, P.:
\newblock Nnvlp: A neural network-based vietnamese language processing toolkit.
\newblock Proceedings of the IJCNLP 2017, System Demonstrations (2017)  37--40

\bibitem{florian2003named}
Florian, R., Ittycheriah, A., Jing, H., Zhang, T.:
\newblock Named entity recognition through classifier combination.
\newblock In: Proceedings of the seventh conference on Natural language
  learning at HLT-NAACL 2003-Volume 4, Association for Computational
  Linguistics (2003)  168--171

\bibitem{TACL792}
Chiu, J., Nichols, E.:
\newblock Named entity recognition with bidirectional lstm-cnns.
\newblock Transactions of the Association for Computational Linguistics
  \textbf{4} (2016)  357--370

\bibitem{Pham2017a}
Pham, T.H., Le, H.P.:
\newblock The importance of automatic syntactic features in vietnamese named
  entity recognition.
\newblock CoRR \textbf{abs/1705.10610} (2017)

\bibitem{Brown:1992:CNG:176313.176316}
Brown, P.F., deSouza, P.V., Mercer, R.L., Pietra, V.J.D., Lai, J.C.:
\newblock Class-based n-gram models of natural language.
\newblock Comput. Linguist. \textbf{18} (1992)  467--479

\bibitem{miller-guinness-zamanian:2004:HLTNAACL}
Miller, S., Guinness, J., Zamanian, A.:
\newblock Name tagging with word clusters and discriminative training.
\newblock In Susan~Dumais, D.M., Roukos, S., eds.: HLT-NAACL 2004: Main
  Proceedings, Boston, Massachusetts, USA, Association for Computational
  Linguistics (2004)  337--342

\bibitem{koo-carreras-collins:2008:ACLMain}
Koo, T., Carreras, X., Collins, M.:
\newblock Simple semi-supervised dependency parsing.
\newblock In: Proceedings of ACL-08: HLT, Columbus, Ohio, Association for
  Computational Linguistics (2008)  595--603

\bibitem{turian-ratinov-bengio:2010:ACL}
Turian, J., Ratinov, L.A., Bengio, Y.:
\newblock Word representations: A simple and general method for semi-supervised
  learning.
\newblock In: Proceedings of the 48th Annual Meeting of the Association for
  Computational Linguistics, Uppsala, Sweden, Association for Computational
  Linguistics (2010)  384--394

\bibitem{liang2005semi}
Liang, P.:
\newblock Semi-supervised learning for natural language.
\newblock PhD thesis, Massachusetts Institute of Technology (2005)

\bibitem{le2017empirical}
Le-Hong, P., Pham, Q.N.M., Pham, T.H., Tran, T.A., Nguyen, D.M.:
\newblock An empirical study of discriminative sequence labeling models for
  vietnamese text processing.
\newblock In: Proceedings of the 9th International Conference on Knowledge and
  Systems Engineering (KSE 2017). (2017)

\bibitem{pennington2014glove}
Pennington, J., Socher, R., Manning, C.D.:
\newblock Glove: Global vectors for word representation.
\newblock In: Empirical Methods in Natural Language Processing (EMNLP). (2014)
  1532--1543

\bibitem{CRFsuite}
Okazaki, N.:
\newblock Crfsuite: a fast implementation of conditional random fields (crfs)
  (2007)

\bibitem{NguyenVNDJ-ALTA-2017}
Nguyen, D.Q., Vu, T., Nguyen, D.Q., Dras, M., Johnson, M.:
\newblock {From Word Segmentation to POS Tagging for Vietnamese}.
\newblock In: Proceedings of the Australasian Language Technology Association
  Workshop 2017. (2017)

\bibitem{nguyen2016hybrid}
Nguyen, T.P., Le, A.C.:
\newblock A hybrid approach to vietnamese word segmentation.
\newblock In: Computing \& Communication Technologies, Research, Innovation,
  and Vision for the Future (RIVF), 2016 IEEE RIVF International Conference on,
  IEEE (2016)  114--119

\bibitem{NguyenNVDJ2018}
Nguyen, D.Q., Nguyen, D.Q., Vu, T., Dras, M., Johnson, M.:
\newblock {A Fast and Accurate Vietnamese Word Segmenter}.
\newblock In: Proceedings of the 11th International Conference on Language
  Resources and Evaluation (LREC 2018). (2018)

\end{thebibliography}
\end{document}